\documentclass[10pt, a4paper]{article}

\usepackage[final]{lrec-coling2024} 
\usepackage{makecell}
\usepackage{multirow}
\usepackage{dialogue}
\usepackage{amssymb}
\usepackage{amsmath}
\usepackage[nolist]{acronym}

\title{Towards a Zero-Data, Controllable, Adaptive Dialog System}

\name{Dirk Väth, Lindsey Vanderlyn, Ngoc Thang Vu} 

\address{University of Stuttgart \\
         Stuttgart, Germany \\
         \{vaethdk|vanderly|thangvu\}@ims.uni-stuttgart.de}

\abstract{
Conversational Tree Search \cite{vath-etal-2023-conversational} is a recent approach to controllable dialog systems, where domain experts shape the behavior of a Reinforcement Learning agent through a dialog tree.
The agent learns to efficiently navigate this tree, while adapting to information needs, e.g., domain familiarity, of different users.
However, the need for additional training data hinders deployment in new domains.
To address this, we explore approaches to generate this data directly from dialog trees.
We improve the original approach, and show that agents trained on synthetic data can achieve comparable dialog success to models trained on human data, both when using a commercial Large Language Model for generation, or when using a smaller open-source model, running on a single GPU.
We further demonstrate the scalability of our approach by collecting and testing on two new datasets: \emph{ONBOARD}, a new domain helping foreign residents moving to a new city, and the medical domain \emph{DIAGNOSE}, a subset of Wikipedia articles related to scalp and head symptoms. 
Finally, we perform human testing, where no statistically significant differences were found in either objective or subjective measures between models trained on human and generated data.
 \\ \newline  
}

\begin{document}

\begin{acronym}
    \acro{LLM}[LLM]{Large Language Model}
    \acro{RL}[RL]{Reinforcement Learning}
    \acro{QA}[QA]{Question Answering}
    \acro{CTS}[CTS]{Conversational Tree Search}
    \acro{NER}[NER]{Named Entity Recognition}
\end{acronym}

\maketitleabstract

\section{Introduction}
While the breakthroughs of modern \acp{LLM} have made the creation of new dialog systems much easier, controlling their generated output remains an open challenge.
This makes \acp{LLM} especially unsuitable for sensitive domains, e.g., legal or medical domains, where users must be able to implicitly trust the system's output.
In such domains, dialog designers usually have the choice between implementing an FAQ-retrieval system or a hand-crafted dialog system. 

FAQ systems directly match user queries to question/answer pairs curated by domain experts, allowing close control of outputted texts \citep{10.1145/1066078.1066079}.
However, as they are single-turn systems and cannot ask clarifying questions, they are only able provide general answers, rather than personalized content for a specific user and their situation.
Including information for multiple cases in one answer would make them unapproachably long, while adding FAQs for each case, would make retrieval challenging.
Retrieval accuracy itself is an open challenge \citep{thakur2021beir},
creating a trade-off: Either providing a single, possibly incorrect answer to a user's question,
or providing multiple answers and shifting the burden of selecting the correct one to the user,
which might be challenging for users unfamiliar with the domain. 

Dialog systems, in contrast, allow for turn-based interactions, which can provide shorter, personalized answers, as well as support users new to a domain without enough experience to formulate precise questions. 
However, such systems either suffer from longer interactions (for handcrafted systems),
or require large amounts of training data \citep{raghu-etal-2021-end} and lack transparency and controllability \citep{10.1145/3209978.3210183} (in the case of machine learning approaches),
making them less suitable for low-resource settings \citep{zhang2020recent} or sensitive domains \citep{cohen2020back}. 

\citet{vath-etal-2023-conversational} address this problem by proposing a new type of hybrid dialog task bridging these two interaction styles, called \ac{CTS}.
In this task, dialog experts first define a dialog tree.
An agent then learns to either walk the user through each node in the tree, or to skip over parts not required to answer a user's more specific question.
In this way, the agent is able to adapt its behavior to the user's preferred interaction style,
supporting both specific and vague user queries,
without sacrificing the controllability required in sensitive domains.

However, \ac{CTS} still requires that dialog designers collect a corpus of real-user utterances, which poses a barrier to scaling this approach to new domains, especially for large and complicated domains.
The goal of this paper is to remove this barrier by exploring how \ac{CTS} can scale to new domains through the use of synthetically generated training data.

Concretely, we seek to answer the following research questions:
\begin{itemize}
    \item \textbf{(RQ1)} How can we effectively generate data for a zero data approach to training \ac{CTS} agents?
        \begin{itemize}
            \item (\textbf{RQ1.1}) How can we analyze the quality of generated data?
            \item (\textbf{RQ1.2}) How do agents trained on generated data perform in simulation, compared to agents trained on human data?
           \item (\textbf{RQ1.3}) How well do the data generation techniques transfer to new domains?
        \end{itemize}
  \item (\textbf{RQ2}) How does a \ac{CTS} agent trained on generated data perform with real users compared to an agent trained on human data?           
\end{itemize}

To address these questions, we investigate how \acp{LLM} can be leveraged to automatically generate training data for new domains,
while at the same maintaining the controllability aspect of the \ac{CTS} task.
We compare the quality of different data generation schemes by evaluating the performance of \ac{RL} agents trained on the synthetic data.
Then, we test scalability of our approach to new domains in simulation using multiple generative \acp{LLM}.
Finally, we perform user testing to verify the transferability to real-world use cases. All code and data is publicly available.\footnote{\url{https://github.com/DigitalPhonetics/conversational-tree-search/tree/generated_v3}}

Our main contributions are: 
1) Creating two new datasets, \emph{ONBOARD} and \emph{DIAGNOSE}.
2) Improving the training procedure for the \ac{CTS} agent, increasing absolute dialog success by more than $18\%$.
3) Introducing a new prompting method for generating diverse data, and demonstrating that automatic diversity and answerability metrics can provide insights for downstream dialog performance.
4) Demonstrating that our generation techniques scale to new domains, where agents trained on synthetic data show comparable (no statistically significant difference) or better dialog success than agents trained on human data. 
5) Showing that success of agents in simulation translates to successful interactions with real users, with no statistically significant differences.

\section{Related Work}

\subsection{Task-oriented Dialog Systems}
While open-domain dialog systems allow users to freely talk about any topic  without a concrete goal, task-oriented dialog systems focus on helping a user reach a specific goal.
Many task-oriented dialog systems use a slot-filling approach, where the dialog system tries to fill values for a selection of slots, e.g. cuisine type, that are necessary to reach that goal from the user \cite{GUS}.
While slot filling approaches can allow hand-crafted dialog policies to follow pre-defined dialog flows \cite{lucas2000voicexml}, or can help efficiently narrowing down searches across e.g. database rows, such as finding restaurants or getting trip recommendations \cite{louvan-magnini-2020-recent}, they are usually unable to perform semantic searches over the dialog domain and in cases of learned systems, unable to follow a dialog-designer controlled flow.

\subsection{Adaptive Dialog Systems}
Research into adaptive dialog systems aims to better align dialog system output with user expectations.
Much research in this area uses generative models to adapt linguistic style,
e.g., adjusting utterances depending on users' emotional states \cite{MA202050} or personalities \citep{doi:10.1137/1.9781611975321.71, 9801557}.
However, generative models are by their nature difficult to control \citep{dusek-kasner-2020-evaluating}.
Some approaches even adapt the complexity of language \citep{janarthanam-lemon-2014-adaptive}.
In order to adapt underlying system behavior, however, additional cues have usually been required, e.g. social cues like  laughter \citep{ritschel2018shaping}, or explicit fine-tuning by the user \cite{chen2012critiquing, narducci2018improving}.
However, eliciting such social cues is difficult for text-based systems and asking for explicit feedback places extra burden on the user.

\subsection{Controllable Dialog Systems}
In sensitive domains, it is crucial subject-experts maintain control of dialog flow to ensure correctness of system outputs.
However, purely handcrafted systems struggle to handle the breadth of possible user inputs.
To this end, several hybrid approaches have been investigated.
Early approaches involved hand-crafting the set of actions  allowed at a given dialog turn \cite{williams2008integrating}.
More recent approaches expand on this idea for neural systems \citep{williams-etal-2017-hybrid, liang2018hierarchical, razumovskaia2019incorporating}, where the action space can be constrained using masks, e.g., by automatically converting expert designed dialog trees into hybrid code networks \citep{shukla2020conversation}.
While such approaches help control dialog agent behavior, they do not provide a mechanism for skipping portions of a dialog irrelevant to a user, which leads to longer interactions that can be frustrating for users with more domain familiarity.

\subsection{Data Generation and Augmentation}

Common data augmentation approaches include lexical substitution \cite{wei-zou-2019-eda}, where tokens are inserted, deleted or substituted with semantically similar replacements, as well as back-translation \cite{sennrich-etal-2016-improving} where data is automatically translated into other languages before being translated back to the source language.
While such approaches can help to expand an existing dataset, they still require seed data, which may not exist for new domains. 

To address this, research in, e.g., the field of low-resource \ac{QA} has started exploring the role of \acp{LLM} in data generation \cite{puri-etal-2020-training,chen2023gotta}.
Given a text, \acp{LLM} can be prompted to generate questions about it, e.g., by asking the model to generate a question for which a given named entity is the answer \citep{li2023selfprompting}.

However, \acp{LLM} are black-box algorithms and suffer from hallucination \cite{azaria2023internal,peng2023check,manakul2023selfcheckgpt}. 
As such, it is difficult to guarantee that the generated questions are logical, natural, or answerable by the original text.
Moreover, commonly used automatic evaluation metrics for text generation do not necessarily correlate with human judgment \cite{nema-khapra-2018-towards}.
In light of this, we explore different generation strategies and techniques for analyzing the artificial data quality, rather than trusting a single metric.

A recent approach in the dialog community trains a model for generating synthetic dialog acts and user utterances for flowchart-grounded troubleshooting dialogs \cite{zhan-etal-2023-turning}.
While this method also relies on the domain representation in form of a structured graph, our generation approach does not require any model training, nor any training data besides the domain graph itself.
Additionally, \ac{CTS} is not limited to the specific task format of trouble-shooting dialogs.

\subsection{Conversational Tree Search}
The goal of \ac{CTS}, as outlined by \citet{vath-etal-2023-conversational}, is to train an \ac{RL} agent to traverse a dialog tree, guiding a user to the answer for a given question.
By using fixed system outputs (which can be personalized via a template mechanism), and by preventing skipping between branches of the dialog tree, the \ac{CTS} task allows subject-experts to maintain controllability.

\begin{figure}[tb]
    \centering
    \includegraphics[width=0.49\textwidth]{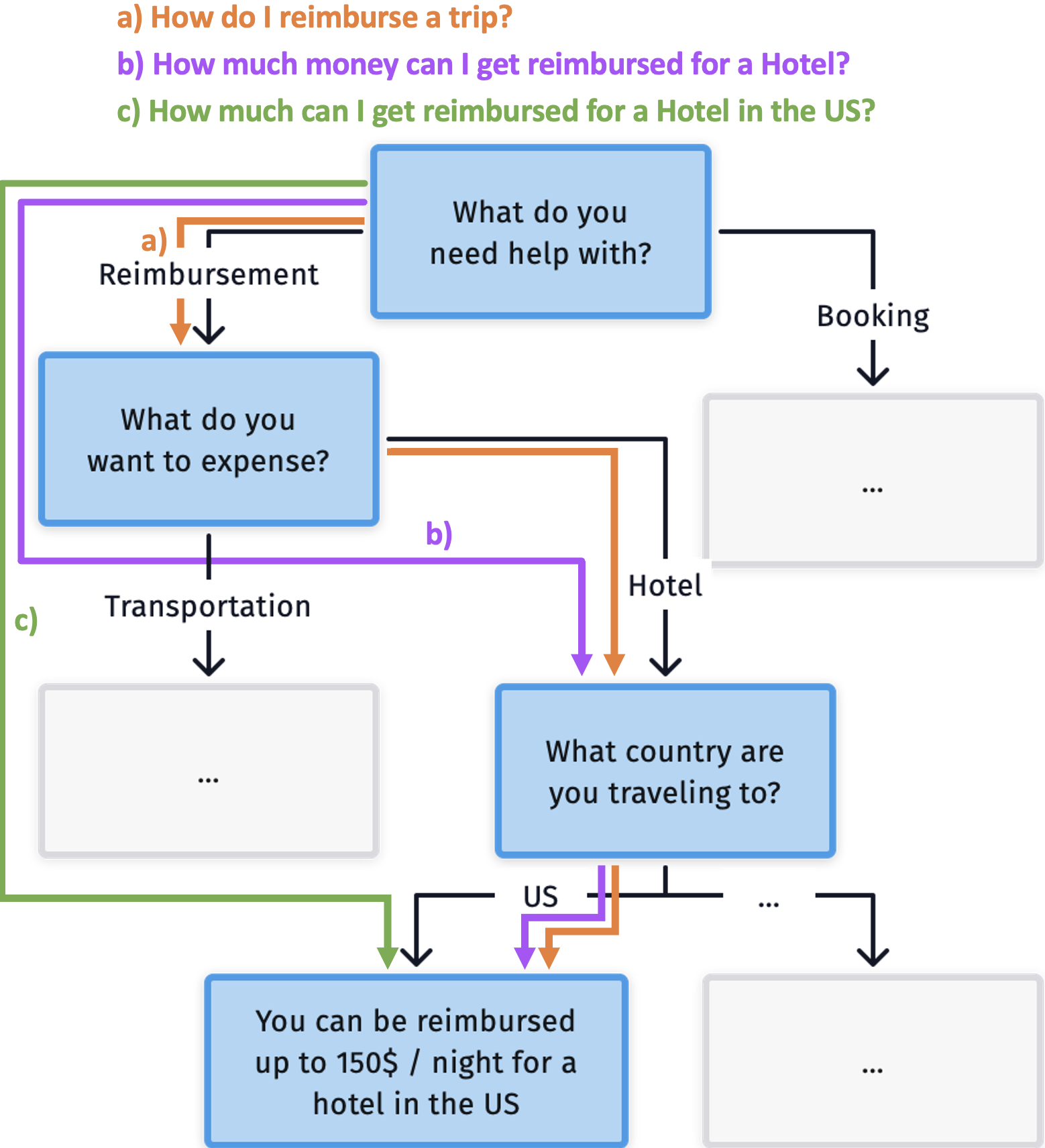}
    \caption{Example of the \ac{CTS} agent adapting its behavior based on the information content of the initial user utterance \cite{vath-etal-2023-conversational}.}
    \label{fig:CTS_example_dialog}
\end{figure}

At the same time, the trained agent can adapt its behavior to different interaction styles, based on the users' utterances.
\ac{CTS} proposes two sub-tasks: guided mode and free mode, representing the extreme cases of information seeking scenarios, as well as the interpolation between.
Guided mode supports users unable to formulate their information need as a specific question, by guiding them step-by-step through each node in the dialog graph (e.g., new users not familiar with a domain).
In contrast, free mode aims to support users with a specific question by learning to skip over as many nodes as possible, while still clarifying the information need enough to deliver an appropriate and personalized answer.
\autoref{fig:CTS_example_dialog} shows three example dialogs for the same user goal, and how a \ac{CTS} agent would adapt to each scenario, deciding to output or skip nodes as needed.

Training is performed against a simulated user, which represents the \ac{RL} environment.
For each simulated dialog, a random goal node is drawn which the simulated user is trying to reach, by asking questions or responding to system requests.

\section{Datasets}
\label{sec:datasets}
To investigate the scalability of our data generation techniques, we examine the performance  of the \ac{CTS} agent on three new datasets, and compare to the original \emph{REIMBURSE} dataset from \citet{vath-etal-2023-conversational}.
In contrast to the \emph{REIMBURSE} dataset, the goal of all new datasets is to serve as a zero-data test-bed for testing training and testing models on data generated directly from the nodes themselves.
While we do provide a test and a train set, like that in \emph{REIMBURSE}, the goal of this is to allow for the training of reference models to act as a benchmark for models trained entirely on generated data.

\begin{table*}[htb]
    \center
    \resizebox{\textwidth}{!}{
        \begin{tabular}{|c|c|c|c|c|c|c|c|c|}
            \hline
            \textbf{Dataset} & \textbf{Split} & \textbf{\#Nodes} & \textbf{\makecell{Tree\\Depth}} & \textbf{\makecell{Max. Node\\Degree}} & \textbf{\makecell{\#User\\Questions}} & \textbf{\makecell{Avg. User\\Questions}} &\textbf{ \makecell{\#Answer\\Paraphrases}} & \textbf{\makecell{Avg. Answer\\Paraphrases}}  \\ 
            \hline
            \multirow{2}{*}{\textbf{REIMBURSE}} & Train & \multirow{2}{*}{123} & \multirow{2}{*}{32} & \multirow{2}{*}{14} & 279 & 3.5 & 246 & 3.4 \\
             & Test &  &  & & 173 & 2.2 & 162 & 2.2 \\ 
            \hline
            \multirow{2}{*}{\textbf{REIMBURSE-En}} & Train & \multirow{2}{*}{123} & \multirow{2}{*}{32} & \multirow{2}{*}{14} & 279 & 3.5 & 246 & 3.4 \\
             & Test & & & & 173 & 2.2 & 162 & 2.2 \\
            \hline
            \textbf{DIAGNOSE} & Train & \multirow{2}{*}{98} & \multirow{2}{*}{10} & \multirow{2}{*}{6} & 219 & 2.9 & 298 & 3.0 \\
            \textbf{DIAGNOSE} & Test & & & & 150 & 2.0 & 298 & 3.0 \\
            \hline
            \textbf{ONBOARD} & Train & \multirow{2}{*}{88}  & \multirow{2}{*}{15} & \multirow{2}{*}{9} & 141 & 2.4 & 175 & 3.1 \\
            \textbf{ONBOARD} & Test & & & & 117 & 2.0 & 152 & 2.7 \\ 
            \hline
        \end{tabular}
    }
    \caption{Overview of original \emph{REIMBURSE}, translated \emph{REIMBURSE-En}, and newly created \emph{ONBOARD} and \emph{DIAGNOSE} datasets (numbers rounded to one decimal).}
    \label{tab:datasets}
\end{table*}

\subsection{REIMBURSE}
The \emph{REIMBURSE} dataset as proposed by \citet{vath-etal-2023-conversational} is a German language dataset for the \ac{CTS} task.
It is a challenging real-world dataset in the travel reimbursement domain, created with domain experts.
Along with the dialog tree, questions and answer paraphrases were collected from real user interactions.
These questions and answer-paraphrases have been split into a train and test set which can each be used by the provided user simulator to generate an arbitrary number of simulated dialogs.
A breakdown of the dataset statistics can be found in \autoref{tab:datasets}.

Although we do not train any new models on this dataset, we use it as a benchmark to compare the performance of our agents to.

\subsection{REIMBURSE-En}
In order to make the \ac{CTS} task more accessible to a wider audience, we choose to translate the \emph{REIMBURSE} dataset to English.
Additionally, this opens up more options for language models and resources, which might not have been available for the original German data.
This dataset represents a direct translation of the \emph{REIMBURSE} dataset, sharing all of the same characteristics, in order to allow for comparisons to the findings of the original \ac{CTS} paper. 
The translation was performed manually by a bilingual domain-expert in order to obtain a faithful and factually correct English equivalent.
Dataset statistics are shown in \autoref{tab:datasets}.

\subsection{DIAGNOSE}

The \emph{DIAGNOSE} dataset was created for the medical domain.
It was designed to help users identify different medical conditions based on symptoms, as well as to find out more about treatment options and risk factors.
The dataset is based on a small subset of Wikipedia articles about conditions related to scalp and head symptoms.
\emph{DIAGNOSE} was designed to be comparatively easy.
Even though the node texts contain a large amount of domain-specific vocabulary, the dialog tree has a lower maximum node degree and a shallower tree depth than \emph{REIMBURSE-En}.
Additionally, the dialog graph for this domain does not contain any variable- or logic nodes.
A breakdown of dataset properties can be found in \autoref{tab:datasets}.

An example node and associated questions can be seen below:

\begin{dialogue}
    \speak{\textbf{Node Text}} Anemia symptoms include fatigue, pale skin and gums, blue color in the whites of the eyes, brittle nails, irritability, dizziness, sore tongue, shortness of breath, unusual food cravings, and headache.
    \speak{\textbf{Question 1}} What are symptoms of anemia?
    \speak{\textbf{Question 2}} How do I know if I have anemia?
    \speak{\textbf{Question 3}} Is a sore tongue a common symptom of anemia?    
\end{dialogue}

\subsection{ONBOARD}

The \emph{ONBOARD} dataset provides users with information about moving to a new city in a foreign country, and the legal and financial steps they will need to undertake, i.e., setting up bank accounts, acquiring health insurance, applying for required visas or residence permits, etc.
This domain presents an additional challenge as it contains code-switching for topics related to legal issues, in order provide users with official names for documents, concepts, and institutions.
Similar to the \emph{REIMBURSE} dataset, the dialog tree for \emph{ONBOARD} contains multiple variable nodes and several logic nodes.
A breakdown of the dataset statistics can be found in \autoref{tab:datasets}.

An example of a a dialog node and test questions is given below.

\begin{dialogue}
    \speak{\textbf{Node Text}} The registration office will provide you with a confirmation of your registration [Meldebest\"atigung], which you will need for opening a bank account and for obtaining a residence permit (if applicable).
    \speak{\textbf{Question 1}} Where do I get confirmation that I've registered my address?
    \speak{\textbf{Question 2}} What do I need the confirmation of registering my address for?
\end{dialogue}

\section{Dialog Agent Implementation}
\label{sec:methods:algorithm}

For our \ac{RL} dialog agent, we follow the architecture and training process outlined in \cite{vath-etal-2023-conversational} with the following changes: 

1) We swap the original language model for an MPNET \cite{song2020mpnet} based Sentence-Transformer \cite{sentencetransformers}, as the new datasets we introduce are in English, and it reports the highest average performance of pretrained Sentence-Transformers for English.

2)  In contrast to free mode, rewards for guided mode only considered whether the agent moved to the correct next node, rather than checking that a global goal was reached by the end of the dialog. 
After analyzing conversations between CTS agent and user simulator obtained by the original implementation, we believe it is more realistic that, even in guided mode, users would have a consistent question they wanted answered. 
Therefore, we now draw global goals for guided mode users (a node anywhere in the graph) instead of choosing one of the immediate neighboring nodes as the next goal each turn.
We then assign a large reward to reaching the global goal.
At the same time, we keep a small positive reward for skipping to the correct follow-up node along the sampled trajectory, as a sequence of locally correct decisions (reaching a correct immediate neighbor) implies global correctness (reaching the correct goal node).
These changes result in a harsher evaluation metric for dialog success, since e.g. in a 5-step dialog, following a correct trajectory, but missing the final goal in the last turn, will now result in a failed dialog ($0\%$ success) instead of a partially successful dialog ($80\%$ success), which we consider to be more realistic.

3) Finally, the original CTS agent was trained jointly on navigating the graph and on predicting the appropriate interaction style (intent). 
Here, we scale the loss of the interaction style prediction objective down to $0.1$ to emphasize learning Q-values as the main task:
$\mathcal{L} = \mathcal{L}_{\text{ddqn}} + 0.1 \mathcal{L}_{\text{intent}}$.
We found this had no significant impact on the interaction style prediction F1 score.

4) We tune several other hyperparameters, increasing the batch size from $128$ to $256$, and the training steps from $1.5e6$ to $2e6$. 

All hyperparameters for training the dialog agent are listed in \autoref{appendix:rlparams}.

\section{Data Generation Methods}

As the user simulator from \citet{vath-etal-2023-conversational} requires both, initial user questions and per-node user responses, we explore methods for generating both of these types of utterances.
We test these generation methods with a small \ac{LLM}, and with a large commercial one, both of which can process separate system and user input directives.

\subsection{Question Generation}
\label{sec:methods:question_generation}

\paragraph{Method 1}
The first method, Gen$_{V1}$, is a naive prompt instructing an \ac{LLM} to generate diverse, FAQ-style questions about a given dialog node's text via the system directive. 
The amount of questions to generate and the node context are then given via user input (see \autoref{tab:q_prompts}).

\paragraph{Method 2} 
For Gen$_{V2}$, we use the same user input, but change the system directive to explicitly generate shorter questions (\autoref{tab:q_prompts}).

\paragraph{Method 3}
For the last method, Gen$_{V3}$, we were inspired by \citet{li2023selfprompting} and \citet{chen2023gotta}, who use \ac{NER} to steer question generation.
However, these approaches only generate cloze questions, where the named entity is the answer, severely limiting the diversity of generated questions \citep{puri-etal-2020-training}.
Therefore, we develop a novel mixed method to increase question diversity.
We first generate 3 questions about the whole node text using the Method 2, to get a basic coverage of the node.
Then, we perform \ac{NER} and explicitly prompt the \ac{LLM} to generate three questions about each entity --instead of forcing the entities to only be the answer-- using a second set of prompts (see \autoref{tab:q_prompts}).
If the total number of generated questions is lower than 10, we generate the difference using Method 2.

\begin{table*}[htb]
    \center
    \resizebox{\textwidth}{!}{
        \begin{tabular}{|c|c|c|l|}
            \hline
            \textbf{Method} & \textbf{Role} & \textbf{Context} & \textbf{\makecell{Prompt}} \\
            \hline
            \multirow{2}{*}{V1} & System & Node text & \makecell[l]{You are a truthful assistant, generating diverse FAQ-style questions given some facts.\\The generated questions should be answerable using the given fact only, without\\additional knowledge. The questions should also be human-like. Try to vary the\\amount of information between questions. Present the results in a numbered list.} \\ \cline{2-4}
            & User & Node Text & Generate 10 FAQ-style questions about the given facts: ``\{NODE TEXT\}''. \\
            \hline

            \multirow{2}{*}{V2} & System & Node Text & \makecell[l]{
            You are a truthful assistant, generating diverse FAQ-style questions given some facts.\\The generated questions should be answerable using the given fact only, without\\additional knowledge. The questions should also be short and human-like. Try to vary\\the amount of information between questions. Present the results in a numbered list.} \\ \cline{2-4}
            & User & Node Text & (same as V1) \\
            \hline

            \multirow{2}{*}{V3} & System & \makecell{Node Text} & (same as V2) \\ \cline{2-4}
            & User & \makecell{Node Text, NER} & \makecell[l]{Generate 3 questions about the entity ``\{NER\}'' from the fact: ``\{NODE TEXT\}''} \\
            \hline
        \end{tabular}
    }
    \caption{Prompt templates for generating synthetic question data.}
    \label{tab:q_prompts}
\end{table*}

\subsection{Response Generation}
\label{sec:methods:answer_generation}

To generate responses, we extract all nodes requiring user input from the dialog graph.
Then, we instruct the \acp{LLM} to generate 5 paraphrases for each possible answer prototype, in the context of the full node text (\autoref{tab:a_prompts}; A).
Additionally, to mimic different user interaction styles, we instruct the \acp{LLM} to generate 5 paraphrases of the the responses using only keywords (\autoref{tab:a_prompts}; B).

\begin{table*}[htb]
    \center
    \resizebox{\textwidth}{!}{
        \begin{tabular}{|c|c|c|l|}
            \hline
            \textbf{Method} & \textbf{Role} & \textbf{Context} & \textbf{\makecell{Prompt}} \\
            \hline
            \multirow{2}{*}{A} & System & Node text & \makecell[l]{You are generating semantically similar paraphrases for a given response to some\\question. The generated response paraphrases should be human-like and short, using \\frequently used words and phrases only. Present the results in a numbered list.} \\ \cline{2-4}
            & User & Node Text & \makecell[l]{Generate 5 paraphrases for the response ``\{RESPONSE TEXT\}'' to the question\\``\{NODE TEXT\}''} \\
            \hline
            
            \multirow{2}{*}{B} & System & \makecell{Node Text} & \makecell[l]{You are shortening a given response to some question into a keyword-like prompt.\\Present the results in a numbered list.} \\ \cline{2-4}
            & User & \makecell{Node Text, NER} & \makecell[l]{
            Generate 5 options for shortening the response ``\{RESPONSE TEXT\}'' to the question\\``\{NODE TEXT\}''
            } \\
            \hline
        \end{tabular}
    }
    \caption{Example of prompting method for generating synthetic user response data.}
    \label{tab:a_prompts}
\end{table*}

\section{Experimental Setup}
\subsection{RQ 1.1: Analysis of Generated Data}
We generate data using the methods described in sections \ref{sec:methods:question_generation} and \ref{sec:methods:answer_generation}.
We use two different \acp{LLM}: ChatGPT (gpt-3.5-turbo, via API) \footnote{\url{https://platform.openai.com/docs/models/gpt-3-5}} and a LLAMA-based \cite{touvron2023llama}, instruction fine-tuned and quantized model \footnote{\url{https://huggingface.co/TheBloke/upstage-llama-30b-instruct-2048-GPTQ}} that fits onto a single NVIDIA GeForce RTX 3090 graphics card. Gen$_{V3}$ uses Stanza \cite{qi2020stanza} for \ac{NER}.

To calculate question similarity, we use the Sentence-Transformer model from section \ref{sec:methods:algorithm}.
Answer confidence scores are calculated with a \ac{QA} model \footnote{\url{https://huggingface.co/deepset/roberta-large-squad2}} pretrained on the SQUAD2.0 dataset \cite{rajpurkar-etal-2018-know}, using a generated question and associated node text that is supposed to contain the answer as inputs.
Finally, we measure diversity using Self-BLEU \cite{selfbleu} scores.

\subsection{RQ 1.2: Human Data vs. Synthetic Data}
\label{exp:evaluation_simulation}
For automatic evaluation, we use the updated \ac{CTS} user simulator (\autoref{sec:methods:algorithm}) with 500 randomly chosen dialog goals on the \emph{REIMBURSE-En} test split.
We evaluate not only the combined success rate (average between guided and free mode success), but also present a metric representing the user's \emph{perceived dialog length}, which counts only the nodes shown to the user.

\subsection{RQ 1.3: Method Generalizability}
To evaluate how well our data generation method generalizes to new domains, we perform additional evaluation in simulation, analogous to (section \ref{exp:evaluation_simulation}), using the test splits of the new datasets \emph{ONBOARD} and \emph{DIAGNOSE}.

\subsection{Human Evaluation (RQ 2)}
To understand how performance of an agent trained on generated data translates to real-world users, we recruit 44 participants from the crowdsourcing platform Prolific\footnote{\url{https://www.prolific.com}} to take part in human evaluation.
Participants were native English speakers with varying experience with business travel (self-rating between 2 and 5 on a 5 point Likert-scale).
They were compensated at the platform recommended rate of 9£/hour.
The experiment took roughly 20 minutes.

\paragraph{Study Design}
We asked each participant to interact with either a CTS agent trained on real data or one trained on generated data in the \emph{REIMBURSE} domain.
Apart from demographic information, we ask for previous experience with dialog systems and with business travel.
During the experiment, participants were asked to complete three conversations with their assigned dialog system.
Each conversation, they were randomly assigned a new goal, covering one of three expected interaction styles: 1) ``open'' goals representing a general/vague information need, 2) ``easy'' goals representing a concrete information need, and 3) ``hard'' goals representing a concrete information need requiring personalized information to correctly answer.
Personalized information refers to the user's specific circumstances, e.g. trip duration or funding organization, which can change the dialog flow.
Between each dialog, users were asked to rate their subjective perception of dialog length and how well their question was answered.
After the interaction, they were asked rate the usability of the dialog agent, how much they trusted it, and its reliability.
For more details see \autoref{appendix:user_study}.

\subsubsection{Evaluation Metrics}

The perceived  dialog length was measured on a 5-point scale from 1 (much too short) to 5 (much too long).
Perceived success was measured on a 4-point scale, where users were asked to rate how well their question had been answered from 1 (not at all) to 4 (completely).
Additionally, the objective dialog length and success condition were logged for each dialog.
Usability of the dialog agent was measured using the Universal Measure of User Experience scale developed by \citet{finstad2010usability}.
User trust was measured using the reliability and trust subscales from \citet{korber2018theoretical}).

\section{Results \& Discussion}

Before testing performance of agents trained on generated data, we first verify our changes to the \ac{CTS} agent.
As the hyperparameters for the original agent were tuned on the German dataset, for fairness, we report the original \ac{CTS} agent's performance on both  English and German (\autoref{tab:agent_reimburse}).

Our changes to the \ac{CTS} agent improve the combined success rate by over $10\%$ compared to the original agent on the German \emph{REIMBURSE} dataset and $18\%$ for the English \emph{REIMBURSE-En}.
It should be noted that the actual improvement over the German agent is likely larger, as the success metric reported for German comes from \citep{vath-etal-2023-conversational}, rather than the new and harsher metric we use for English (\autoref{sec:methods:algorithm}).

\subsection{RQ 1: Transitioning to a Zero Data Approach}

\paragraph{RQ 1.1 Analyzing the quality of generated data}

\begin{figure}[htb]
    \centering
    \includegraphics[width=0.49\textwidth]{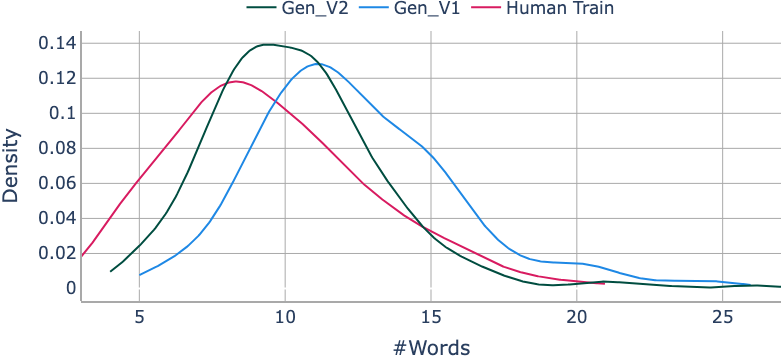}
    \caption{Smoothed density plot of question lengths from human data and generated data.}
    \label{fig:gen_question_lengths}
\end{figure}

Looking at the question lengths between human data and data generated by \textit{GEN$_{V1}$} (\autoref{fig:gen_question_lengths}),
we observe that the generated questions seem to be longer than human questions.
When manually inspecting the generated questions, we also find them to be much less natural than those from the human data.

We amend the original prompt used, creating \textit{GEN$_{V2}$}, to explicitly ask for short outputs (\autoref{sec:methods:question_generation}) in an effort to align the syntax of the generations better with the human data.
This change to the prompt shifts the distribution of question lengths more towards the human training distribution, and qualitatively yields more natural utterances.
However, it still does not ensure that the artificial data is semantically similar to human data.

\begin{figure}[t]
    \centering
    \includegraphics[width=0.49\textwidth]{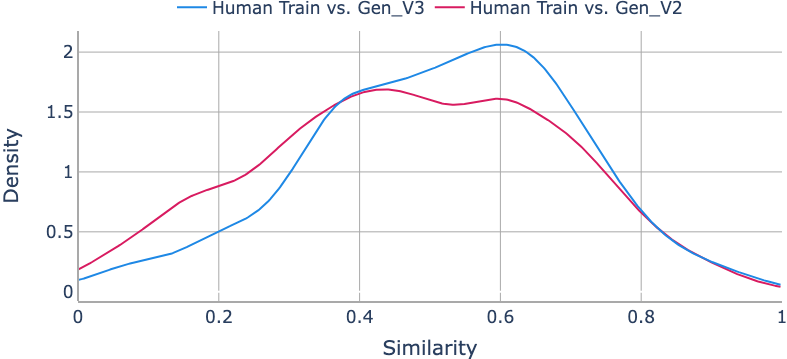}
    \caption{Smoothed density plot of question similarities between human generated training data.}
    \label{fig:gen_question_similarities}
\end{figure}

To investigate how semantically similar the generated questions are to human data, we calculate the pair-wise similarities between all human and generated questions for each node from the dialog graph, and then average the similarities across all nodes (\autoref{fig:gen_question_similarities}). 
Here, we see that the Gen$_{V2}$ data is still quite distinct from the human data.

When manually inspecting the generations, we find that generated questions tend to focus only on one part of the node text, making them lack diversity and omit topics real users might ask about.
To address this, we develop the novel two-step Gen$_{V3}$ prompt, steering the model to explicitly ask about all named entities in a node (\autoref{sec:methods:question_generation}).
We see that doing so significantly ($p < 1.82e-11$) increases the similarity of the generated (avg.: $0.52$) to the human training data than Gen$_{V2}$ (avg.: $0.47$), as measured with a standard t-test.

\begin{table}[b]
    \center
    \resizebox{0.45\textwidth}{!}{
        \begin{tabular}{|c|c|c|c|c|c|}
            \hline
            \makecell{Training Data} & n-1 & n-2 & n-3 & n-4 & n-5 \\
            \hline
            Human & \textbf{0.78} & \textbf{0.68} & \textbf{0.60} & \textbf{0.54} & \textbf{0.49} \\ \hline
            V1 & 0.95 & 0.92 & 0.87 & 0.83 & 0.80 \\
            V2 & 0.95 & 0.90 & 0.85 & 0.80 & 0.76 \\
            V3 & \textbf{0.85} & \textbf{0.78} & \textbf{0.71} & \textbf{0.66}& \textbf{0.62} \\
            \hline
        \end{tabular}
    }
    \caption{Self-BLEU scores for different n-gram sizes on human and generated data.}
    \label{tab:my_label}
\end{table}

We also look at the diversity of the generated questions.
The self-BLEU scores (\autoref{tab:my_label}) show that the Gen$_{V3}$ data are the most diverse.
This metric can be used to analyze the quality of the generated data even in the absence of human comparison data.

\begin{table*}[h!t]
    \center
    \resizebox{\textwidth}{!}{
        \begin{tabular}{|l|c|c|c|c|c|c|}
            \hline
             \textbf{Model} & \textbf{\makecell{Training\\Data}} & \textbf{\makecell{Avg. Perceived\\Length (guided)}} & \textbf{\makecell{Avg. Perceived\\Length (free)}} & \textbf{\makecell{Success\\(combined)}} & \textbf{\makecell{Dialog Mode\\Prediction F1}} & \textbf{\makecell{Dialog Mode\\Prediction Consistency}} \\
             \hline
             Original & human (GER) & n/a & n/a & 62.58\% & 0.85 & \textbf{1.0} \\
             Original & human (EN) & n/a & n/a & 55.28\% & 0.86 & 0.87 \\
             Ours & human (EN)
             & 13.56 & 2.95 & \textbf{73.86\%} & \textbf{0.94} & 0.96 \\
             \hline
             Ours & V1 (LLAMA) & 13.53 & 3.41 & 64.17\% & \textbf{0.98} & \textbf{0.97} \\
             Ours & V2 (LLAMA) & 11.71 & 3.65 & 65.02\% & \textbf{0.98} & 0.95 \\
             Ours & V3 (LLAMA) & 12.89 & 3.45 & \textbf{69.44\%} & 0.96 & 0.95 \\
             \hline
             Ours & V1 (ChatGPT) & 13.02 & 3.65 & 64.35\% & \textbf{0.98} & \textbf{0.97}\\
             Ours & V2 (ChatGPT) & 14.55 & 3.71 & 66.67\% & 0.95  & \textbf{0.97}  \\
             Ours & V3 (ChatGPT) & 12.87 & 3.59 & \textbf{68.41}\% & \textbf{0.98} & \textbf{0.97}  \\
             \hline
        \end{tabular}
    }
    \caption{Simulation results on \emph{REIMBURSE(-En)} test splits of original \ac{CTS} agent (German), our improved agent (English), and our \ac{CTS} agent trained on generated data only (English).
    }
    \label{tab:agent_reimburse}
\end{table*}

In conjunction with diversity, we estimate the average ``answerability'' via \ac{QA} confidence scores of the generated questions, given the node text as answer.
Here, we also see that the improvements from Gen$_{V3}$ and Gen$_{V2}$ together also significantly ($p < 0.0003$) increase the average answerability, from an average of $0.36$ with naive prompt to $0.42$ with Gen$_{V3}$, according to a t-test.

When looking at downstream performance (\autoref{tab:agent_reimburse}), we see that improvements in these metrics also lead to higher dialog success, suggesting they can be used as an indicator of generation quality.

\paragraph{RQ1.2: Human Data vs. Synthetic Data}

To investigate whether synthetic data can be a viable alternative to human data, we compare agent performance in simulation.
From \autoref{tab:agent_reimburse}, we see that the best performing agent trained on artificial data (Gen$_{V3}$: 69.44\% success) performs comparably to the best performing agent trained on human data (CTS$_{ours}$: 73.86\% success).
Using a standard t-test, we find no statistically significant difference.

\paragraph{RQ1.3: Generalizing to new domains}
To test of the scalability of our generation methods, we analyze model performance on two new domains. 
As each of these has their own challenges (\autoref{sec:datasets}), we compare each model trained on generated data to a baseline trained on human data.

\begin{table*}[htb]
    \center
    \resizebox{\textwidth}{!}{
        \begin{tabular}{|c|c|c|c|c|}
            \hline
            \textbf{Domain} & \textbf{Training Data} & \textbf{Avg. Perceived Length (guided)} & \textbf{Avg. Perceived Length (free)} & \textbf{\makecell{Success (combined)}} \\
            \hline

            DIAGNOSE & human         & 6.42 & 2.29 & 76.31\% \\
            DIAGNOSE & V3 (LLAMA)    & 6.62 & 2.95 & 71.08\% \\
            DIAGNOSE & V3 (ChatGPT)  & 5.65 & 2.46 & \textbf{85.12\%} \\
            \hline

            ONBOARD & human        & 7.88 & 2.98 & \textbf{73.61\%} \\
            ONBOARD & V3 (LLAMA)   & 7.91 & 3.52 & 63.38\% \\      
            ONBOARD & V3 (ChatGPT) & 7.60 & 3.58 & 70.72\% \\
            \hline
        \end{tabular}
    }
    \caption{Performance of \ac{CTS} agents trained on human and generated data on the new domains \emph{DIAGNOSE} and \emph{ONBOARD} in simulation.}
    \label{tab:domain_generalization}
\end{table*}

When looking at \autoref{tab:domain_generalization}, the agent trained on data generated by LLAMA is again nearly able to match the performance of the model trained on human data for the \emph{DIAGNOSE} dataset, while the model trained on data generated by ChatGPT surpasses it.
On the other hand, the \emph{ONBOARD} dataset may present a more challenging domain, due in part to the code-switching present in the dialog nodes. 
Despite this, the model trained on data from ChatGPT nearly reaches the performance of models trained on human data.

Based on this, we find that the generation techniques do appear to scale to new domains, as t-tests show no statistically significant differences between the best synthetically trained agents and the agents trained on real data in any domain.

\subsection{RQ 2: Human Evaluation}

\subsubsection{Generated vs. Real Data}
\label{sec:human_obj}

After performing human evaluation, we find that there are no statistically significant differences (using a standard t-test) between either subjective or objective measures of success or dialog length (\autoref{tab:joint_data}).
Additionally, we find no difference in the reported trust, reliability, or usability scores between either group.
This suggests that there is no human-observable loss in performance when using generated data compared to real data, either in terms of objective metrics or subjective metrics.

\begin{table}[htb]
    \center
    \resizebox{0.49\textwidth}{!}{
         \begin{tabular}{ |c|c|c|c|c| } 
         \hline
          \textbf{\makecell{Training\\Data}} & 
          \textbf{\# Turns}  & 
          \textbf{Success} & 
          \makecell{\textbf{Perceived} \\ \textbf{Length}} & 
          \makecell{\textbf{Answer} \\ \textbf{Satisfaction}} 
          \\ \hline

         Human  & 6.14  & 77.59  & 2.88 &  2.93  \\
         V3  & 5.27  & 72.73  & 2.65 &  2.73\\
         \hline
        \end{tabular}
    }
    \caption{Average objective and subjective performance metrics of a \ac{CTS} agent trained on human data vs. generated data.}
    \label{tab:joint_data}
\end{table}

\subsubsection{Human Evaluation vs. Simulator}
Finally, to validate our updated user simulator, we additionally compare the objective performance metrics from the human evaluation (\autoref{tab:joint_data}) to those obtained in simulation (\autoref{tab:agent_reimburse}).
We find that the success rates between the simulated and human dialogs are very comparable ($73.86\%$ and $77.59\%$ respectively for the model trained on human data, and $69.44\%$ and $72.73\%$ for the model trained on generated data).
We perform statistical analysis using Welch's t-test to account for the difference in sample size, and find no significant difference, regardless of the source of training data.

Based on this, we conclude that results from simulation translate well to real human interaction, suggesting the simulator can be a good proxy for real user evaluation.
We therefore expect the results reported in (\autoref{tab:domain_generalization}) will translate to similar performance with real users.

\section{Conclusion}

In this paper, we present two new and publicly available datasets, \emph{ONBOARD}, providing help for moving to a new city in a foreign country, and \emph{DIAGNOSE}, a medical domain. 
The datasets each consist of a dialog tree and human-collected text inputs.

We apply a harsher, more realistic evaluation metric and improve on the agent training method from the original \ac{CTS} \citep{vath-etal-2023-conversational}, increasing dialog success by over $18\%$.

Given a dialog tree, we explore several zero-data prompting-based methods for generating user utterance data to train a \ac{CTS} agent, developing a novel two-stage prompting approach to increase question diversity.
Through this process, we find that automatic scores for diversity and answerability can be indicative of downstream dialog task performance.

Furthermore, we show that there is no statistically significant difference in objective metrics between agents trained on human data or on generated data in the \emph{REIMBURSE-En} domain.
We verify this both through simulation and through testing with real users.
User evaluation further reveals no statistically significant differences on subjective metrics (trust, reliability, usability, subjective length, or subjective dialog success) either.
This suggests that we can effectively generate training data from a dialog tree, such that \ac{CTS} agents can be trained in zero data settings with negligible performance loss.
We also find that the size of the tested \acp{LLM} does not result in significant differences in task performance. 

To evaluate how well our techniques scale to new domains, we further tested agent performance on both new datasets we introduced.
For \emph{ONBOARD}, we again find that performance of agents trained on generated data is comparable to that of agents trained on human data.
For \emph{DIAGNOSE}, performance can even exceed that of the agent trained on human data.
This suggests that our methods scale well to new domains.

\section{Ethical Considerations}

To ensure that users could give informed consent, we provided a detailed description of the task and research objectives both on the crowdsourcing platform and once they had accepted the task.
In respect of participant privacy, we specifically did not collect personally identifying data from any users.
To this end, we store all logs and survey responses using an anonymous hash generated based on a given username, rather than with the username itself.
In this way, users could log in again if they needed to take a break in the middle of the interaction, but we had no way of directly linking any recorded results to, e.g., users' Prolific account identifiers.
To ensure that participants were fairly compensated, we followed best practices recommended by the crowdsourcing platform paying users at 9£/hr.
We additionally used our pilot study to verify that our estimated time was below the median time we selected when advertising the task.

\section{Limitations}
While we try to cover many different real-world use cases with the presented domains, we cannot account for the challenges of all possible future domains.
Additionally, although our work removes the necessity to collect training data, creating a dialog tree is still required (which may be large for complex domains).
Finally, replicating the exact data generated and analyzed in this paper depends on the specific versions of the \acp{LLM} used.

\section{Bibliographical References}\label{sec:reference}

\bibliographystyle{lrec_natbib}
\bibliography{lrec-coling2024-example}

\begin{thebibliography}{41}
\expandafter\ifx\csname natexlab\endcsname\relax\def\natexlab#1{#1}\fi

\bibitem[{Azaria and Mitchell(2023)}]{azaria2023internal}
Amos Azaria and Tom Mitchell. 2023.
\newblock The internal state of an llm knows when its lying.
\newblock \emph{arXiv preprint arXiv:2304.13734}.

\bibitem[{Bobrow et~al.(1977)Bobrow, Kaplan, Kay, Norman, Thompson, and
  Winograd}]{GUS}
Daniel~G. Bobrow, Ronald~M. Kaplan, Martin Kay, Donald~A. Norman, Henry
  Thompson, and Terry Winograd. 1977.
\newblock Gus, a frame-driven dialog system.
\newblock \emph{Artificial Intelligence}.

\bibitem[{Chen and Pu(2012)}]{chen2012critiquing}
Li~Chen and Pearl Pu. 2012.
\newblock Critiquing-based recommenders: survey and emerging trends.
\newblock \emph{User Modeling and User-Adapted Interaction}, 22(1):125--150.

\bibitem[{Chen et~al.(2023)Chen, Zhang, Deng, Jiang, and Wang}]{chen2023gotta}
Xiusi Chen, Yu~Zhang, Jinliang Deng, Jyun-Yu Jiang, and Wei Wang. 2023.
\newblock Gotta: generative few-shot question answering by prompt-based cloze
  data augmentation.
\newblock In \emph{Proceedings of the 2023 SIAM International Conference on
  Data Mining (SDM)}, pages 909--917. SIAM.

\bibitem[{Cohen(2020)}]{cohen2020back}
Philip~R. Cohen. 2020.
\newblock \href {https://ojs.aaai.org/index.php/AAAI/article/view/7073} {Back
  to the future for dialogue research}.
\newblock In \emph{The Thirty-Fourth {AAAI} Conference on Artificial
  Intelligence, {AAAI} 2020, The Thirty-Second Innovative Applications of
  Artificial Intelligence Conference, {IAAI} 2020, The Tenth {AAAI} Symposium
  on Educational Advances in Artificial Intelligence, {EAAI} 2020, New York,
  NY, USA, February 7-12, 2020}, pages 13514--13519. {AAAI} Press.

\bibitem[{Du{\v{s}}ek and Kasner(2020)}]{dusek-kasner-2020-evaluating}
Ond{\v{r}}ej Du{\v{s}}ek and Zden{\v{e}}k Kasner. 2020.
\newblock \href {https://aclanthology.org/2020.inlg-1.19} {Evaluating semantic
  accuracy of data-to-text generation with natural language inference}.
\newblock In \emph{Proceedings of the 13th International Conference on Natural
  Language Generation}, pages 131--137, Dublin, Ireland. Association for
  Computational Linguistics.

\bibitem[{Finstad(2010)}]{finstad2010usability}
Kraig Finstad. 2010.
\newblock The usability metric for user experience.
\newblock \emph{Interacting with computers}, 22(5):323--327.

\bibitem[{Firdaus et~al.(2023)Firdaus, Shandilya, Ekbal, and
  Bhattacharyya}]{9801557}
Mauajama Firdaus, Arunav Shandilya, Asif Ekbal, and Pushpak Bhattacharyya.
  2023.
\newblock \href {https://doi.org/10.1109/TCSS.2022.3182986} {Being polite:
  Modeling politeness variation in a personalized dialog agent}.
\newblock \emph{IEEE Transactions on Computational Social Systems},
  10(4):1455--1464.

\bibitem[{Gao et~al.(2018)Gao, Galley, and Li}]{10.1145/3209978.3210183}
Jianfeng Gao, Michel Galley, and Lihong Li. 2018.
\newblock \href {https://doi.org/10.1145/3209978.3210183} {Neural approaches to
  conversational ai}.
\newblock In \emph{The 41st International ACM SIGIR Conference on Research \&
  Development in Information Retrieval}, SIGIR '18, page 1371–1374, New York,
  NY, USA. Association for Computing Machinery.

\bibitem[{Janarthanam and Lemon(2014)}]{janarthanam-lemon-2014-adaptive}
Srinivasan Janarthanam and Oliver Lemon. 2014.
\newblock \href {https://doi.org/10.1162/COLI_a_00203} {Adaptive generation in
  dialogue systems using dynamic user modeling}.
\newblock \emph{Computational Linguistics}, 40(4):883--920.

\bibitem[{K{\"o}rber(2018)}]{korber2018theoretical}
Moritz K{\"o}rber. 2018.
\newblock Theoretical considerations and development of a questionnaire to
  measure trust in automation.
\newblock In \emph{Congress of the International Ergonomics Association}, pages
  13--30. Springer.

\bibitem[{Li et~al.(2023)Li, Zhang, and Zhao}]{li2023selfprompting}
Junlong Li, Zhuosheng Zhang, and Hai Zhao. 2023.
\newblock \href {http://arxiv.org/abs/2212.08635} {Self-prompting large
  language models for zero-shot open-domain qa}.

\bibitem[{Liang and Yang(2018)}]{liang2018hierarchical}
Weiri Liang and Meng Yang. 2018.
\newblock \href {https://doi.org/10.1007/978-3-319-95933-7\_24} {Hierarchical
  hybrid code networks for task-oriented dialogue}.
\newblock In \emph{Intelligent Computing Theories and Application - 14th
  International Conference, {ICIC} 2018, Wuhan, China, August 15-18, 2018,
  Proceedings, Part {II}}, volume 10955 of \emph{Lecture Notes in Computer
  Science}, pages 194--204. Springer.

\bibitem[{Louvan and Magnini(2020)}]{louvan-magnini-2020-recent}
Samuel Louvan and Bernardo Magnini. 2020.
\newblock \href {https://doi.org/10.18653/v1/2020.coling-main.42} {Recent
  neural methods on slot filling and intent classification for task-oriented
  dialogue systems: A survey}.
\newblock In \emph{Proceedings of the 28th International Conference on
  Computational Linguistics}, pages 480--496, Barcelona, Spain (Online).
  International Committee on Computational Linguistics.

\bibitem[{Lucas(2000)}]{lucas2000voicexml}
Bruce Lucas. 2000.
\newblock Voicexml for web-based distributed conversational applications.
\newblock \emph{Communications of the ACM}, 43(9):53--57.

\bibitem[{Ma et~al.(2020)Ma, Nguyen, Xing, and Cambria}]{MA202050}
Yukun Ma, Khanh~Linh Nguyen, Frank~Z. Xing, and Erik Cambria. 2020.
\newblock \href {https://doi.org/https://doi.org/10.1016/j.inffus.2020.06.011}
  {A survey on empathetic dialogue systems}.
\newblock \emph{Information Fusion}, 64:50--70.

\bibitem[{Manakul et~al.(2023)Manakul, Liusie, and
  Gales}]{manakul2023selfcheckgpt}
Potsawee Manakul, Adian Liusie, and Mark~JF Gales. 2023.
\newblock Selfcheckgpt: Zero-resource black-box hallucination detection for
  generative large language models.
\newblock \emph{arXiv preprint arXiv:2303.08896}.

\bibitem[{Narducci et~al.(2018)Narducci, de~Gemmis, Lops, and
  Semeraro}]{narducci2018improving}
Fedelucio Narducci, Marco de~Gemmis, Pasquale Lops, and Giovanni Semeraro.
  2018.
\newblock Improving the user experience with a conversational recommender
  system.
\newblock In \emph{International Conference of the Italian Association for
  Artificial Intelligence}, pages 528--538. Springer.

\bibitem[{Nema and Khapra(2018)}]{nema-khapra-2018-towards}
Preksha Nema and Mitesh~M. Khapra. 2018.
\newblock \href {https://doi.org/10.18653/v1/D18-1429} {Towards a better metric
  for evaluating question generation systems}.
\newblock In \emph{Proceedings of the 2018 Conference on Empirical Methods in
  Natural Language Processing}, pages 3950--3959, Brussels, Belgium.
  Association for Computational Linguistics.

\bibitem[{Peng et~al.(2023)Peng, Galley, He, Cheng, Xie, Hu, Huang, Liden, Yu,
  Chen et~al.}]{peng2023check}
Baolin Peng, Michel Galley, Pengcheng He, Hao Cheng, Yujia Xie, Yu~Hu, Qiuyuan
  Huang, Lars Liden, Zhou Yu, Weizhu Chen, et~al. 2023.
\newblock Check your facts and try again: Improving large language models with
  external knowledge and automated feedback.
\newblock \emph{arXiv preprint arXiv:2302.12813}.

\bibitem[{Puri et~al.(2020)Puri, Spring, Shoeybi, Patwary, and
  Catanzaro}]{puri-etal-2020-training}
Raul Puri, Ryan Spring, Mohammad Shoeybi, Mostofa Patwary, and Bryan Catanzaro.
  2020.
\newblock \href {https://doi.org/10.18653/v1/2020.emnlp-main.468} {Training
  question answering models from synthetic data}.
\newblock In \emph{Proceedings of the 2020 Conference on Empirical Methods in
  Natural Language Processing (EMNLP)}, pages 5811--5826, Online. Association
  for Computational Linguistics.

\bibitem[{Qi et~al.(2020)Qi, Zhang, Zhang, Bolton, and Manning}]{qi2020stanza}
Peng Qi, Yuhao Zhang, Yuhui Zhang, Jason Bolton, and Christopher~D. Manning.
  2020.
\newblock \href {https://nlp.stanford.edu/pubs/qi2020stanza.pdf} {Stanza: A
  {Python} natural language processing toolkit for many human languages}.
\newblock In \emph{Proceedings of the 58th Annual Meeting of the Association
  for Computational Linguistics: System Demonstrations}.

\bibitem[{Raghu et~al.(2021)Raghu, Agarwal, Joshi, and
  {Mausam}}]{raghu-etal-2021-end}
Dinesh Raghu, Shantanu Agarwal, Sachindra Joshi, and {Mausam}. 2021.
\newblock \href {https://doi.org/10.18653/v1/2021.emnlp-main.357} {End-to-end
  learning of flowchart grounded task-oriented dialogs}.
\newblock In \emph{Proceedings of the 2021 Conference on Empirical Methods in
  Natural Language Processing}, pages 4348--4366, Online and Punta Cana,
  Dominican Republic. Association for Computational Linguistics.

\bibitem[{Rajpurkar et~al.(2018)Rajpurkar, Jia, and
  Liang}]{rajpurkar-etal-2018-know}
Pranav Rajpurkar, Robin Jia, and Percy Liang. 2018.
\newblock \href {https://doi.org/10.18653/v1/P18-2124} {Know what you don{'}t
  know: Unanswerable questions for {SQ}u{AD}}.
\newblock In \emph{Proceedings of the 56th Annual Meeting of the Association
  for Computational Linguistics (Volume 2: Short Papers)}, pages 784--789,
  Melbourne, Australia. Association for Computational Linguistics.

\bibitem[{Razumovskaia and Eskenazi(2019)}]{razumovskaia2019incorporating}
Evgeniia Razumovskaia and Maxine Eskenazi. 2019.
\newblock Incorporating rules into end-to-end dialog systems.
\newblock In \emph{Proc. 3rd NeurIPS Workshop on Conversational AI, Vancouver,
  Canada}, pages 1--11.

\bibitem[{Reimers and Gurevych(2019)}]{sentencetransformers}
Nils Reimers and Iryna Gurevych. 2019.
\newblock \href {https://doi.org/10.18653/v1/D19-1410} {Sentence-bert: Sentence
  embeddings using siamese bert-networks}.
\newblock In \emph{Proceedings of the 2019 Conference on Empirical Methods in
  Natural Language Processing and the 9th International Joint Conference on
  Natural Language Processing, {EMNLP-IJCNLP} 2019, Hong Kong, China, November
  3-7, 2019}, pages 3980--3990. Association for Computational Linguistics.

\bibitem[{Ritschel and Andr{\'e}(2018)}]{ritschel2018shaping}
Hannes Ritschel and Elisabeth Andr{\'e}. 2018.
\newblock Shaping a social robot’s humor with natural language generation and
  socially-aware reinforcement learning.
\newblock In \emph{Proceedings of the workshop on NLG for human--robot
  interaction}, pages 12--16.

\bibitem[{Sennrich et~al.(2016)Sennrich, Haddow, and
  Birch}]{sennrich-etal-2016-improving}
Rico Sennrich, Barry Haddow, and Alexandra Birch. 2016.
\newblock \href {https://doi.org/10.18653/v1/P16-1009} {Improving neural
  machine translation models with monolingual data}.
\newblock In \emph{Proceedings of the 54th Annual Meeting of the Association
  for Computational Linguistics (Volume 1: Long Papers)}, pages 86--96, Berlin,
  Germany. Association for Computational Linguistics.

\bibitem[{Shukla et~al.(2020)Shukla, Liden, Shayandeh, Kamal, Li, Mazzola,
  Park, Peng, and Gao}]{shukla2020conversation}
Swadheen Shukla, Lars Liden, Shahin Shayandeh, Eslam Kamal, Jinchao Li, Matt
  Mazzola, Thomas Park, Baolin Peng, and Jianfeng Gao. 2020.
\newblock \href {https://doi.org/10.18653/v1/2020.acl-demos.39} {{C}onversation
  {L}earner - a machine teaching tool for building dialog managers for
  task-oriented dialog systems}.
\newblock In \emph{Proceedings of the 58th Annual Meeting of the Association
  for Computational Linguistics: System Demonstrations}, pages 343--349,
  Online. Association for Computational Linguistics.

\bibitem[{Song et~al.(2020)Song, Tan, Qin, Lu, and Liu}]{song2020mpnet}
Kaitao Song, Xu~Tan, Tao Qin, Jianfeng Lu, and Tie-Yan Liu. 2020.
\newblock Mpnet: Masked and permuted pre-training for language understanding.
\newblock \emph{Advances in Neural Information Processing Systems},
  33:16857--16867.

\bibitem[{Thakur et~al.(2021)Thakur, Reimers, R{\"u}ckl{\'e}, Srivastava, and
  Gurevych}]{thakur2021beir}
Nandan Thakur, Nils Reimers, Andreas R{\"u}ckl{\'e}, Abhishek Srivastava, and
  Iryna Gurevych. 2021.
\newblock \href
  {https://datasets-benchmarks-proceedings.neurips.cc/paper/2021/file/65b9eea6e1cc6bb9f0cd2a47751a186f-Paper-round2.pdf}
  {Beir: A heterogenous benchmark for zero-shot evaluation of information
  retrieval models}.
\newblock In \emph{35th Conference on Neural Information Processing Systems
  (NeurIPS 2021) Track on Datasets and Benchmarks}.

\bibitem[{Touvron et~al.(2023)Touvron, Lavril, Izacard, Martinet, Lachaux,
  Lacroix, Rozi{\`e}re, Goyal, Hambro, Azhar et~al.}]{touvron2023llama}
Hugo Touvron, Thibaut Lavril, Gautier Izacard, Xavier Martinet, Marie-Anne
  Lachaux, Timoth{\'e}e Lacroix, Baptiste Rozi{\`e}re, Naman Goyal, Eric
  Hambro, Faisal Azhar, et~al. 2023.
\newblock Llama: Open and efficient foundation language models.
\newblock \emph{arXiv preprint arXiv:2302.13971}.

\bibitem[{V{\"a}th et~al.(2023)V{\"a}th, Vanderlyn, and
  Vu}]{vath-etal-2023-conversational}
Dirk V{\"a}th, Lindsey Vanderlyn, and Ngoc~Thang Vu. 2023.
\newblock \href {https://doi.org/10.18653/v1/2023.eacl-main.91} {Conversational
  tree search: A new hybrid dialog task}.
\newblock In \emph{Proceedings of the 17th Conference of the European Chapter
  of the Association for Computational Linguistics}, pages 1264--1280,
  Dubrovnik, Croatia. Association for Computational Linguistics.

\bibitem[{Wei and Zou(2019)}]{wei-zou-2019-eda}
Jason Wei and Kai Zou. 2019.
\newblock \href {https://doi.org/10.18653/v1/D19-1670} {{EDA}: Easy data
  augmentation techniques for boosting performance on text classification
  tasks}.
\newblock In \emph{Proceedings of the 2019 Conference on Empirical Methods in
  Natural Language Processing and the 9th International Joint Conference on
  Natural Language Processing (EMNLP-IJCNLP)}, pages 6382--6388, Hong Kong,
  China. Association for Computational Linguistics.

\bibitem[{Williams(2008)}]{williams2008integrating}
Jason~D Williams. 2008.
\newblock Integrating expert knowledge into pomdp optimization for spoken
  dialog systems.
\newblock In \emph{Proceedings of the AAAI-08 Workshop on Advancements in POMDP
  Solvers}, volume~2, page~25.

\bibitem[{Williams et~al.(2017)Williams, Asadi, and
  Zweig}]{williams-etal-2017-hybrid}
Jason~D. Williams, Kavosh Asadi, and Geoffrey Zweig. 2017.
\newblock \href {https://doi.org/10.18653/v1/P17-1062} {Hybrid code networks:
  practical and efficient end-to-end dialog control with supervised and
  reinforcement learning}.
\newblock In \emph{Proceedings of the 55th Annual Meeting of the Association
  for Computational Linguistics (Volume 1: Long Papers)}, pages 665--677,
  Vancouver, Canada. Association for Computational Linguistics.

\bibitem[{Wu et~al.(2005)Wu, Yeh, and Chen}]{10.1145/1066078.1066079}
Chung-Hsien Wu, Jui-Feng Yeh, and Ming-Jun Chen. 2005.
\newblock \href {https://doi.org/10.1145/1066078.1066079} {Domain-specific faq
  retrieval using independent aspects}.
\newblock \emph{ACM Transactions on Asian Language Information Processing},
  4(1):1–17.

\bibitem[{Yang et~al.(2018)Yang, Qu, Lei, Zhu, Zhao, Chen, and
  Huang}]{doi:10.1137/1.9781611975321.71}
Min Yang, Qiang Qu, Kai Lei, Jia Zhu, Zhou Zhao, Xiaojun Chen, and Joshua~Z.
  Huang. 2018.
\newblock \href {https://doi.org/10.1137/1.9781611975321.71}
  {\emph{Investigating Deep Reinforcement Learning Techniques in Personalized
  Dialogue Generation}}, pages 630--638.

\bibitem[{Zhan et~al.(2023)Zhan, Maruf, Qu, Wang, Zukerman, and
  Haffari}]{zhan-etal-2023-turning}
Haolan Zhan, Sameen Maruf, Lizhen Qu, Yufei Wang, Ingrid Zukerman, and
  Gholamreza Haffari. 2023.
\newblock \href {https://aclanthology.org/2023.alta-1.9} {Turning flowchart
  into dialog: Augmenting flowchart-grounded troubleshooting dialogs via
  synthetic data generation}.
\newblock In \emph{Proceedings of the 21st Annual Workshop of the Australasian
  Language Technology Association}, pages 88--99, Melbourne, Australia.
  Association for Computational Linguistics.

\bibitem[{Zhang et~al.(2020)Zhang, Takanobu, Zhu, Huang, and
  Zhu}]{zhang2020recent}
Zheng Zhang, Ryuichi Takanobu, Qi~Zhu, MinLie Huang, and XiaoYan Zhu. 2020.
\newblock \href {https://doi.org/10.1007/s11431-020-1692-3} {Recent advances
  and challenges in task-oriented dialog systems}.
\newblock \emph{Science China Technological Sciences}, 63(10):2011--2027.

\bibitem[{Zhu et~al.(2018)Zhu, Lu, Zheng, Guo, Zhang, Wang, and Yu}]{selfbleu}
Yaoming Zhu, Sidi Lu, Lei Zheng, Jiaxian Guo, Weinan Zhang, Jun Wang, and Yong
  Yu. 2018.
\newblock \href {https://doi.org/10.1145/3209978.3210080} {Texygen: A
  benchmarking platform for text generation models}.
\newblock In \emph{The 41st International ACM SIGIR Conference on Research \&
  Development in Information Retrieval}, SIGIR '18, page 1097–1100, New York,
  NY, USA. Association for Computing Machinery.

\end{thebibliography}


\newpage
\appendix
\onecolumn

\section{Reinforcement Learning Agent Training Parameters}
\label{appendix:rlparams}

\begin{table}[htb]
    \centering
    \begin{tabular}{c|c}
         \textbf{Parameter} & \textbf{Value} \\ \hline
         Optimizer & Adam \\
         Learning Rate & $1e^{-4}$ \\
         $\lambda$ & $0.1$ \\
         Maximum Training Dialog Turns & $2M$ \\
         Max. Gradient Norm & $1.0$ \\
         Batch Size & $256$ \\
         $\gamma$ & $0.99$ \\
         Exploration fraction of Training Turns & $0.99$ \\
         Exploration Scheme & $\epsilon$-greedy \\
         $\epsilon$ start & $0.6$ \\
         $\epsilon$ end & $0.0$ \\
         Training frequency (w.r.t. dialog turns) & $3$ \\
         Training start (w.r.t. dialog turns) & $1280$ \\
         DDQN Target Network update frequency (w.r.t. training steps) & $15$ \\
         Q-Value clipping & $10.0$ \\
         Munchausen $\tau$ & $0.03$ \\
         Munchausen $\alpha$ & $0.9$ \\
         Munchausen Clipping & $-1$ \\
         Evaluation frequency (w.r.t. dialog turns) & $10000$ \\
         Evaluation dialogs & $500$
    \end{tabular}
    \caption{Hyperparameters for training the Reinforcement Learning agents.}
\end{table}

\newpage
\section{User Study}
\label{appendix:user_study}

\subsection{Data Agreement}
Before beginning the experiment, users were provided with a data agreement. 
Although we did not collect any personally identifying data, we wanted to make sure that users were aware of what they would be asked to do, the purpose of the research, what data we would collect and how the data would be processed. 

\begin{figure}[!htb]
    \centering
    \includegraphics[width=\textwidth]{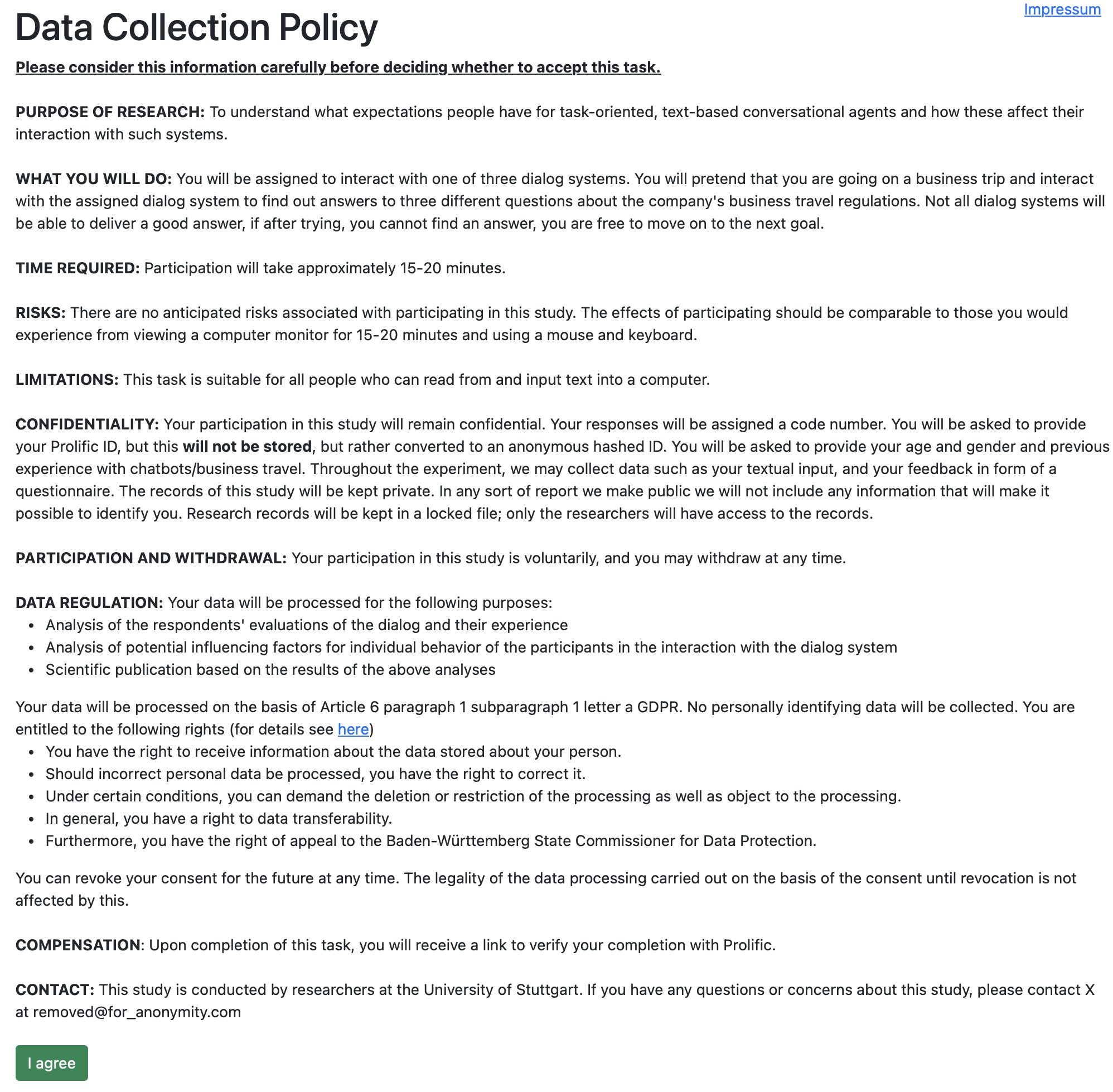}
\end{figure}

\newpage
\subsection{Study Instructions}
During the interaction, users were provided with the following interface, on the right side they had an information goal for which they should find an answer.
On the left side, they had a window with their conversation with the chatbot.
Once they felt they had found an answer to their question, they could click on the button underneath the goal to move on to the next dialog.

\begin{figure}[!htb]
    \centering
    \includegraphics[width=\textwidth]{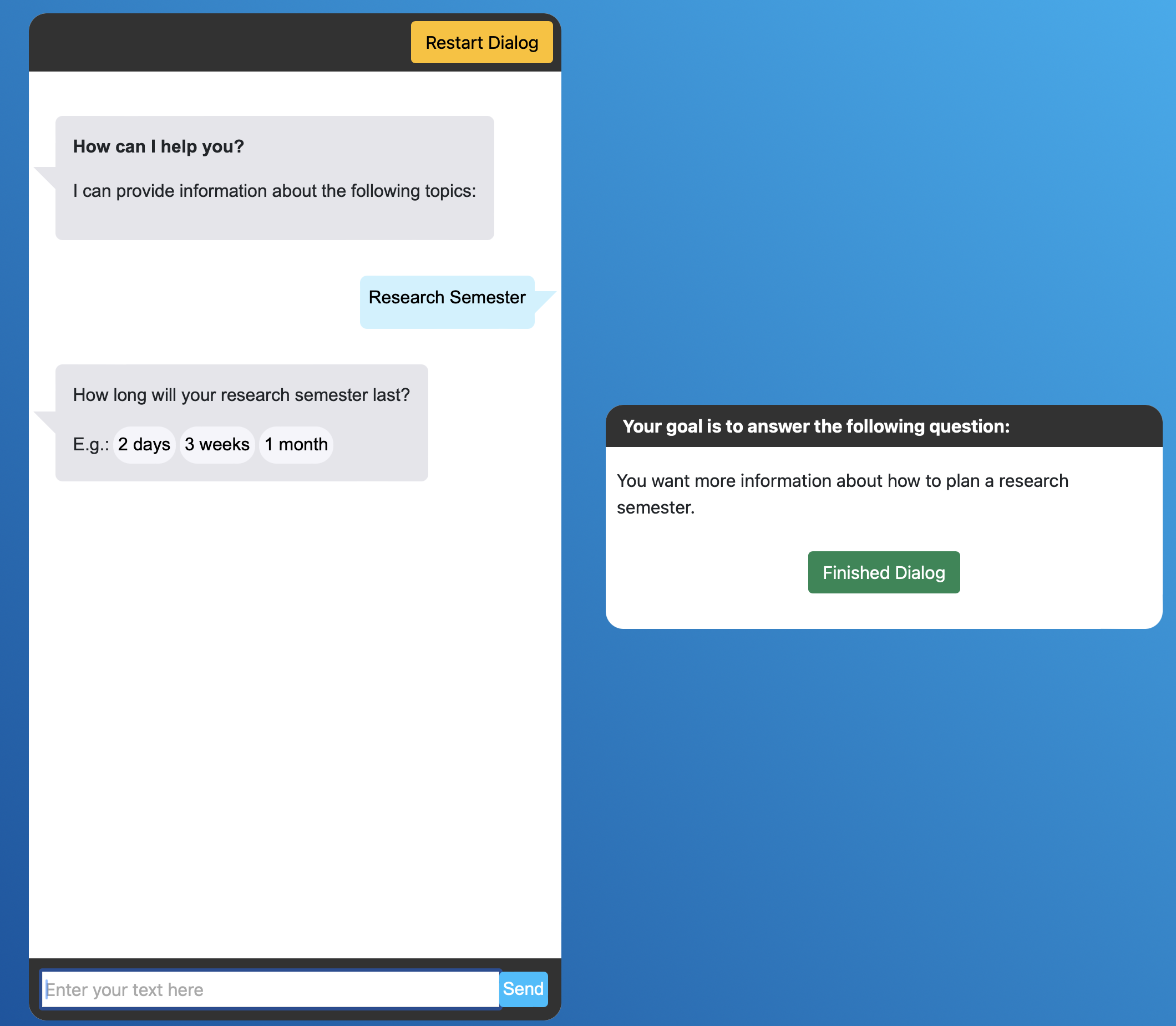}
\end{figure}

\newpage
\subsection{Interaction Surveys}
\label{sec:surveys}

\subsubsection{Pre-Interaction Survey}
The survey given to users before the interaction can be seen below. 
Here they were asked general questions about their demographics, previous experience with the domain and chatbots.

\begin{figure}[!htb]
    \centering
    \includegraphics[width=\textwidth]{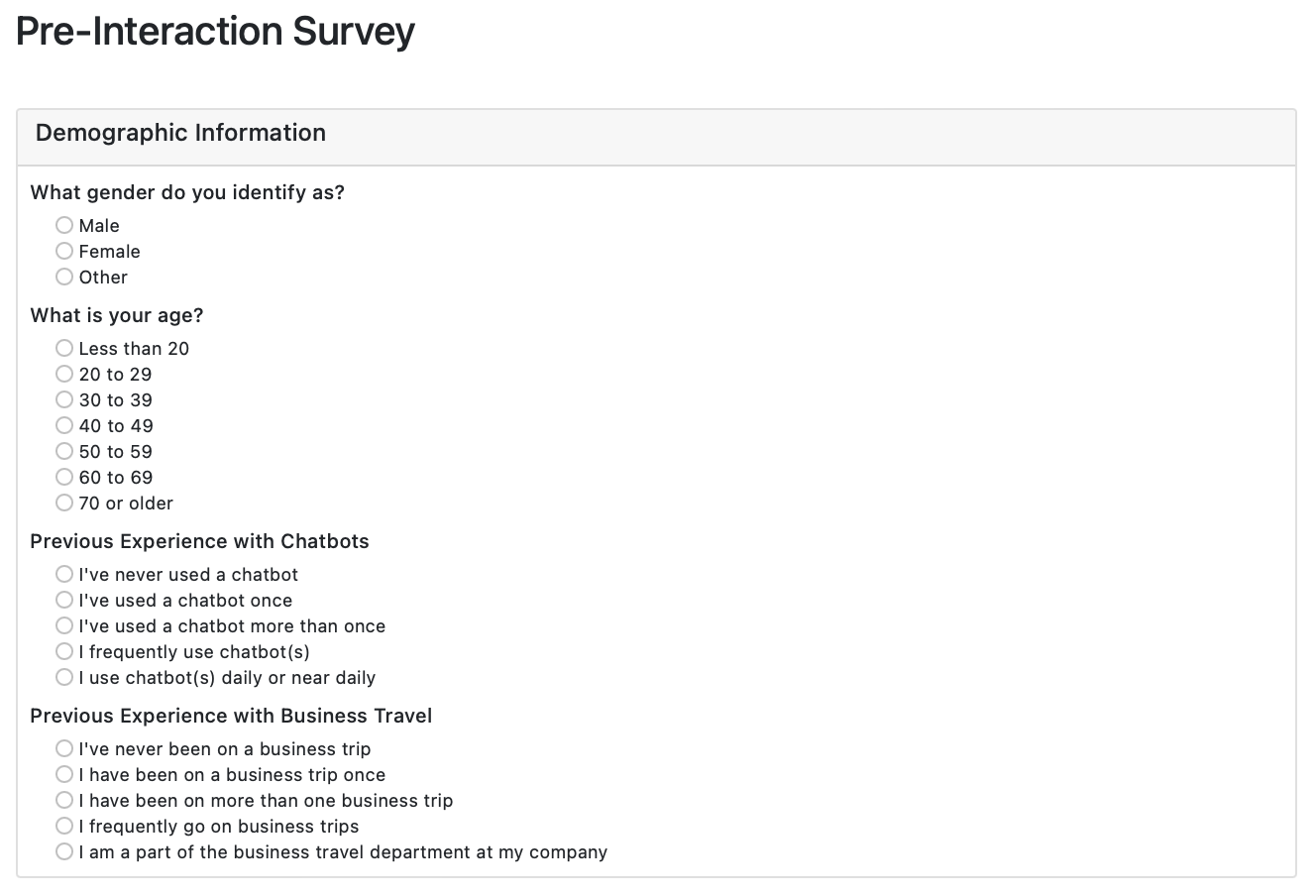}
\end{figure}

\subsubsection{Post-Dialog Survey}

After each interaction, users were asked to rate their perception of the dialog length on a five-point Likert scale and their perception of how well their question was answered on a four-point Likert scale.

\begin{figure}[!htb]
    \centering
    \includegraphics[width=0.75\textwidth]{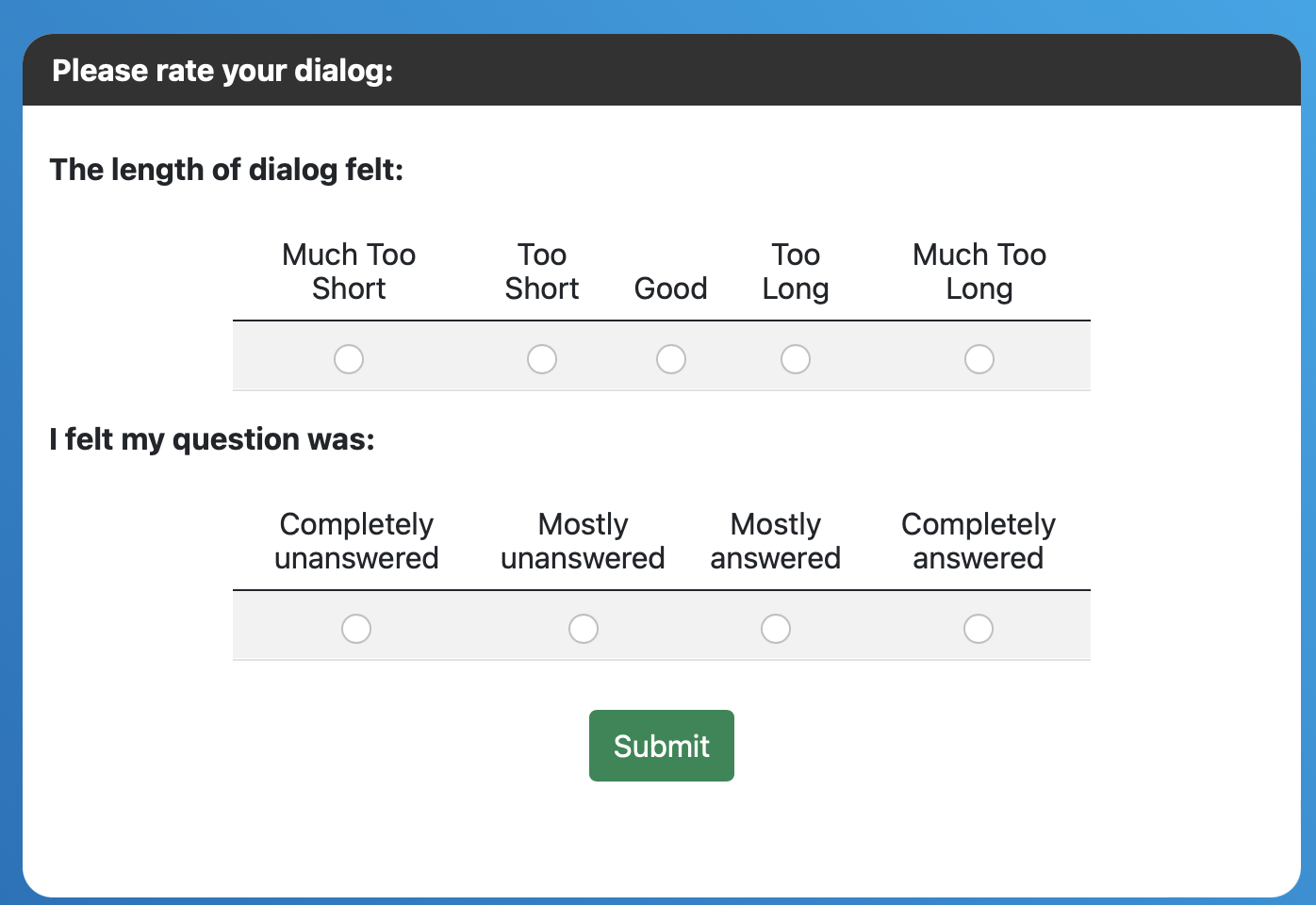}
\end{figure}

\newpage
\subsubsection{Post-Interaction survey}

The survey given to users after interacting with their assigned style of chatbot can be seen below.
Users were asked to fill out a usability questionnaire \citep{finstad2010usability}
and the trust and reliability subscales from the trust in automation questionnaire \citep{korber2018theoretical}.

\begin{figure}[!htb]
    \centering
    \includegraphics[width=\textwidth]{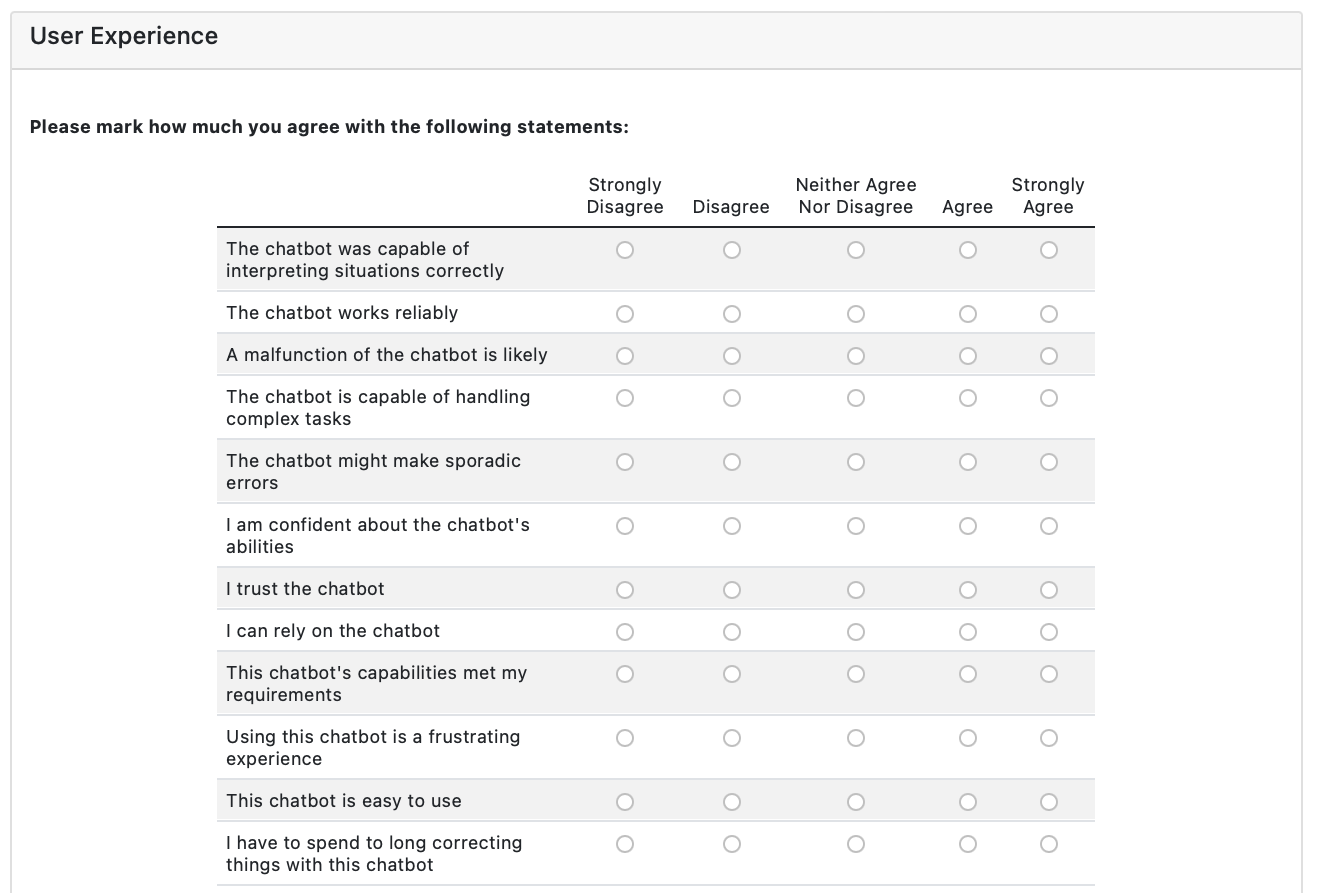}
\end{figure}

\end{document}